\documentclass[runningheads]{llncs}
\usepackage{graphicx}
\usepackage{amsmath,amssymb} % define this before the line numbering.
\usepackage{color}
\usepackage{times}
\usepackage{multirow}
\usepackage{slashbox}

\def\textvspace{{\vspace{-4mm}}}
\def\figvspace{{\vspace{-3mm}}}
\newcommand{\Paragraph}[1]{\vspace{-0mm} \noindent \textbf{#1} \hspace{0mm}}
\newcommand{\Section}[1]{\vspace{-1mm} \section{#1} \vspace{0mm}}
\newcommand{\SubSection}[1]{\vspace{-1mm} \subsection{#1} \vspace{-0mm}}

\begin{document}
% \linenumbers
\pagestyle{headings}
\mainmatter

\title{Face De-Spoofing: Anti-Spoofing via Noise Modeling} 
% Replace with your title

\titlerunning{Face De-Spoofing}
% Replace with a meaningful short version of your title

\author{Amin Jourabloo\thanks{denotes equal contribution by the authors.}, Yaojie Liu$^{\star}$, Xiaoming Liu}

\authorrunning{Amin Jourabloo, Yaojie Liu, Xiaoming Liu}
% Replace with shorter version of the author list. If there are more authors than fits a line, please use A. Author et al.

%Please write out author names in full in the paper, i.e. full given and family names. 
%If any authors have names that can be parsed into FirstName LastName in multiple ways, please include the correct parsing, in a comment to the volume editors:
%\index{Lastnames, Firstnames}
%(Do not uncomment it, because you may introduce extra index items if you do that, we will use scripts for introducing index entries...)

\institute{Department of Computer Science and Engineering,\\
	Michigan State University\\
	\email{ \{jourablo,liuyaoj1,liuxm\}@msu.edu}
}

\maketitle

\begin{abstract}
%With increasing usage of face recognition systems in our life and improving quality of the presentation attacks, face anti-spoofing become a critical step before utilizing face recognition. 
%Previous face anti-spoofing works build discriminative models to recognize the subtle texture differences (i.e., spoof noise) between live and spoof.
Many prior face anti-spoofing works develop discriminative models for recognizing the subtle differences between live and spoof faces.
%Even though previous face anti-spoofing systems have acceptable intra testing performance, they struggle to generalize well for cross testing. 
Those approaches often regard the image as an indivisible unit, and process it holistically, without explicit modeling of the spoofing process.
%due to a lack of study on spoof noise pattern
%In this paper, we propose a new prospective for tackling this problem by inversely decomposing a spoof face image to the live image and the spoof noise pattern without having ground truth of both. 
%In this work, we study several properties of the spoof noise, and propose a new perspective for face anti-spoofing: inversely decomposing a spoof face into the spoof noise and the live face as a denoising process. The estimated spoof noise is inherently discriminative for classification.
In this work, motivated by the noise modeling and denoising algorithms, we identify a new problem of face de-spoofing, for the purpose of anti-spoofing: inversely decomposing a spoof face into a spoof noise and a live face, and then utilizing the spoof noise for classification.
%A novel CNN architecture with multiple loss functions and supervisions is proposed. We encourage the estimated spoof noise pattern to be spatial invariant and repetitive. 
A CNN architecture with proper constraints and supervisions is proposed to overcome the problem of having no ground truth for the decomposition.
We evaluate the proposed method on multiple face anti-spoofing databases. 
The results show promising improvements due to our spoof noise modeling. 
Moreover, %the estimated spoof noise provides an intuitive visual understanding and support of the rationale behind the texture-based face anti-spoofing.    
the estimated spoof noise provides a visualization which helps to understand the added spoof noise by each spoof medium.

\keywords{Face anti-spoofing, Generative model, CNN, Image decomposition}
\end{abstract}

\textvspace
\Section{Introduction}

% Importance of the face anti-spoofing 
With the increasing influence of smart devices in our daily lives, people are seeking for secure and convenient ways to access their personal information. 
Biometrics, such as face, fingerprint, and iris, are widely utilized for person authentication due to their intrinsic distinctiveness and convenience to use. 
Face, as one of the most popular modalities, has received increasing attention in the academia and industry in the recent years (e.g., iPhone X). 
However, the attention also brings a growing incentive for hackers to design biometric presentation attacks (PA), or spoofs, to be authenticated as the genuine user.
%While face recognition can achieve a high accuracy, the quality of the biometric spoofs or Presentation Attacks (PA) are enhancing tremendously. 
Due to the almost no-cost access to the human face, the spoof face can be as simple as a printed photo paper (i.e., print attack) and a digital image/video (i.e., replay attack), or as complicated as a $3$D Mask and facial cosmetic makeup. 
With proper handling, those spoofs can be visually very close to the genuine user's live face.
%Some examples of PA include print, replay, makeup and $3$D mask. 
As a result, these call for the need of developing robust face anti-spoofing algorithms. % in conjunction with the face recognition systems. 

As the most common spoofs, print attack and replay attack have been well studied previously, from different perspectives. 
The cue-based methods aim to detect liveness cues~\cite{pan2007eyeblink,patel2016cross} (e.g., eye blinking, head motion) to classify live videos. But these methods can be fooled by video replay attacks. 
The texture-based methods attempt to compare texture difference between live and spoof faces, using pre-defined features such as LBP~\cite{de2012lbp,de2013can}, HOG~\cite{komulainen2013context,yang2013face}. 
Similar to texture-based methods, CNN-based methods~\cite{li2016original,patel2016cross,yang2014learn} design a unified process of feature extraction and classification. With a softmax loss based binary supervision, they have the risk of overfitting  on the training data. 
%In [], auxiliary supervisions are utilized for reducing the generalization problem of CNN-based methods. 
%Finally, the temporal methods~\cite{agarwal2016face,bao2009liveness,xu2015learning,learning-deep-models-for-face-anti-spoofing-binary-or-auxiliary-supervision} extract features across the video frames. 
Regardless of the perspectives, almost all the prior works treat face anti-spoofing as a {\it black box} binary classification problem.
In contrast, we propose to open the black box by modeling the process of how a spoof image is
generated from its original live image.
%As a generative model being able to provide better representation and generalization than a discriminative model,  in this work, we propose to open the black box by modeling the process of how a spoof image is generated from its original live image.

%In contrast to the previous approaches which consider the face anti-spoofing as a binary classification problem, we propose to model the process of generating the spoof image. Our proposed method maps the spoof image to the corresponding live one by estimation the spoof noise pattern and presents a deeper understanding about the spoof generation process.

%Many image processing tasks [] are designed for recovering the original image which is manipulated in some processes and become noisy. 
\begin{figure*}[t!]
	\centering
	\small
	\includegraphics[width=\linewidth]{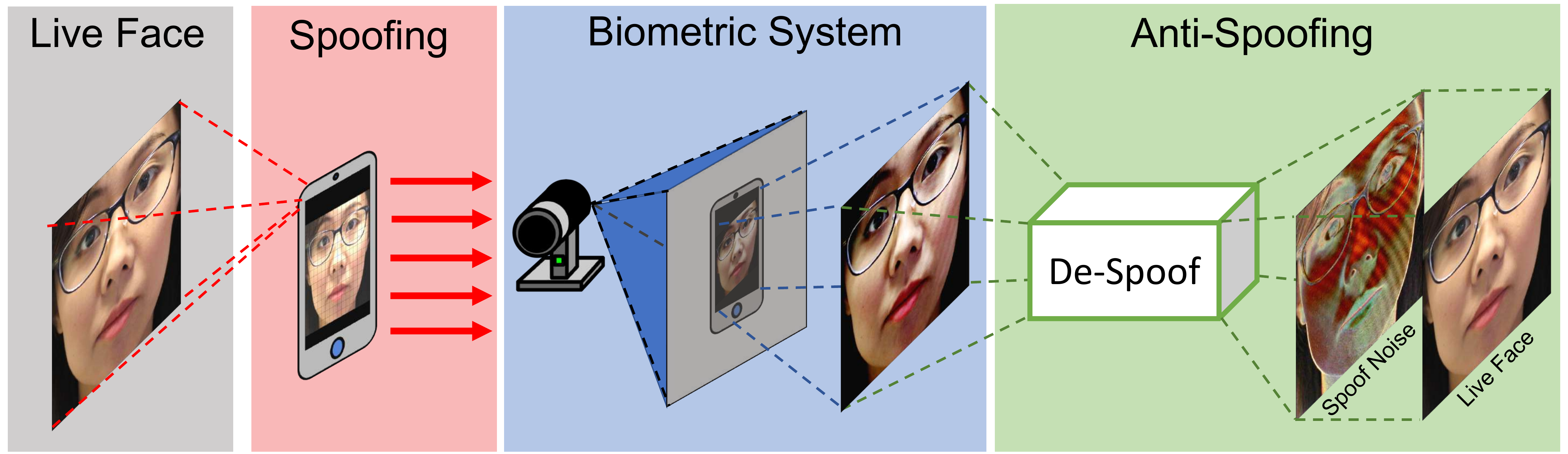} %FIXME change De-Spoof to be De-spoofing.
	\caption{\small The illustration of face spoofing and anti-spoofing processes. De-spoofing process aims to estimate a spoof noise from a spoof face and reconstruct the live face. The estimated spoof noise should be discriminative for face anti-spoofing.} 
	\label{fig:1}
	\figvspace
\end{figure*}
Our approach is motivated by the classic image %restoration 
de-X problems, such as image de-noising and de-blurring~\cite{dong2014learning,jourabloo2012new,kulkarni2016reconnet,pathak2016context}.
In image de-noising, the corrupted image is regarded as a degradation from the additive noise, e.g., salt-and-pepper noise and white Gaussian noise. 
In image de-blurring, the uncorrupted image is degraded by motion, which can be described as a process of convolution.
%in image inpainting~\cite{pathak2016context}, the noise is impacted by the pixel-wise mask of $\mathbf{x}$. 
Similarly, in face anti-spoofing, the spoof image can be viewed as a re-rendering of the live image but with some ``special" noise from the spoof medium and the environment. 
Hence, the natural question is, {\it can we recover the underlying live image 
when given a spoof image, similar to image de-noising}?

Yes. 
This paper shows ``how" to do this.
We call the process of decomposing a spoof face to the spoof noise pattern and a live face as \textit{Face De-spoofing}, shown in Fig.~\ref{fig:1}.
Similar to the previous de-X works, the degraded image $\mathbf{x}\in \mathbb{R}^{m}$ can be formulated as a function of the original image $\mathbf{\hat{x}}$, the degradation matrix $\mathbf{A}\in \mathbb{R}^{m\times m}$ and an additive noise $\mathbf{n}\in \mathbb{R}^{m}$. 
%an image dependent noise by \textit{noise function} $N(.)$,
\begin{equation}
\mathbf{x}=\mathbf{A}\mathbf{\hat{x}} + \mathbf{n} = \mathbf{\hat{x}} + (\mathbf{A}-\mathbb{I}) \mathbf{\hat{x}}  + \mathbf{n} = \mathbf{\hat{x}} +N(\mathbf{\hat{x}}),
\label{eqn_noisemodeling}
\end{equation}
%where the degraded image $\mathbf{x}  \in \mathbb{R}^{m}$ can be considered as a function of the original image $\mathbf{\hat{x}}$, the degradation matrix $\mathbf{A} \in \mathbb{R}^{m\times m}$ and a white Gaussian noise $\mathbf{n}\in \mathbb{R}^{m}$. 
%an image dependent noise by \textit{noise function} $N(.)$,
%\begin{equation}
%\mathbf{x}=\mathbf{A}\mathbf{\hat{x}} + \mathbf{n} = \mathbf{\hat{x}} + (\mathbf{A}-\mathbb{I}) \mathbf{\hat{x}}  + \mathbf{n} = \mathbf{\hat{x}} +N(\mathbf{\hat{x}}),
%\label{eqn_noisemodeling}
%\end{equation}
%this is a linear inverse problem since the noise function can be expressed as, 
%\begin{equation}\label{eq:Model}
%N(\mathbf{\hat{x}})=(\mathbf{A}-\mathbf{I}) \mathbf{\hat{x}},
%\end{equation}
%where $\mathbf{A} \in \mathbb{R}^{m\times m}$ is the manipulative noise operator and $\mathbf{n} \in \mathbb{R}^{m}$ is the white noise. 
%where $\mathbf{A}$ describes the degradation that is dependent to $\mathbf{\hat{x}}$ such as motion blurring, and $\mathbf{n}$ denotes noise that is independent to  $\mathbf{\hat{x}}$. By combining these two terms, $\mathbf{x}$ can be modeled as a summation of the original image $\mathbf{\hat{x}}$ and  the image-dependent noise function $N(\mathbf{\hat{x}})=(\mathbf{A}-\mathbb{I}) \mathbf{\hat{x}}  + \mathbf{n}$. 
where $N(\mathbf{\hat{x}})=(\mathbf{A}-\mathbb{I}) \mathbf{\hat{x}}  + \mathbf{n}$ is the image-dependent noise function.
Instead of solving $\mathbf{A}$ and $\mathbf{n}$, we decide to estimate $N(\mathbf{\hat{x}})$ directly since it is more solvable under the deep learning framework~\cite{lefkimmiatis2017non,taimemnet,tai2017image,zhou2018deep,gharbi2016deep}.
Essentially, by estimating $N(\mathbf{\hat{x}})$ and $\mathbf{\hat{x}}$, we aim to peel off the spoof noise and reconstruct the original live face.
%, as if the spoofing did not happen.
Likewise, if given a live face, face de-spoofing model should return itself plus {\it zero} noise.
Note that our face de-spoofing is designed to handle paper attack, replay attack and possibly make-up attack, but our experiments are limited to the first two PAs.
The benefits of face de-spoofing are twofold: 1) it reverses, or undoes, the spoofing generation process, which helps us to model and visualize the spoof noise pattern of different spoof mediums. 2) the spoof noise itself is discriminative between live and spoof images and hence is useful for face anti-spoofing.

While face de-spoofing shares the same challenges as other image de-X problems, it has a few distinct difficulties to conquer:

\Paragraph{No Ground Truth:}  Image de-X works often use synthetic data where the original undegraded image could be used as ground truth for supervised learning. 
In contrast, we have no access to $\mathbf{\hat{x}}$, which is the corresponding live face of a spoof face image.   

\Paragraph{No Noise Model:} There is no comprehensive study and understanding about the spoof noise. Hence it is not clear how we can constrain the solution space to {\it faithfully} estimate the spoof noise pattern. 

\Paragraph{Diverse Spoof Mediums:} Each type of spoofs utilizes different spoof mediums for generating spoof images. Each spoof medium represents a specific type of noise pattern.

To address these challenges, we propose several constraints and supervisions based on our prior knowledge and the conclusions from a case study (in Section~\ref{case_study}). 
Given that a live face has no spoof noise, we impose the constraint that $N(\mathbf{\hat{x}})$ of a live image is {\it zero}. 
Based on our study,  we assume that the spoof noise of a spoof image is ubiquitous, i.e., it exists everywhere
in the spatial domain of the image; and is repetitive, i.e., it is the spatial repetition of certain noise in the image.
The repetitiveness can be encouraged by maximizing the high-frequency magnitude of the estimated noise in the Fourier domain. 

With such constraints and auxiliary supervisions proposed in~\cite{learning-deep-models-for-face-anti-spoofing-binary-or-auxiliary-supervision}, a novel CNN architecture is presented in this paper. 
Given an image, one CNN is designed to synthesize the spoof noise pattern and reconstruct the corresponding live image.
In order to examine the reconstructed live image, we train another CNN with auxiliary supervision and a GAN-like discriminator in an end-to-end fashion.
These two networks are designed to ensure the quality of the reconstructed image regarding its discriminativeness between live and spoof, and the visual plausibility of the synthesized live image. 

To summarize, the main contributions of this work include:

$\diamond$ We offer a new perspective for detecting the spoofing face from print attack and replay attack by inversely decomposing a spoof face image into the live face and the spoofing noise, without having
the ground truth of either.

$\diamond$ A novel CNN architecture is proposed for face de-spoofing, where appropriate constraints and auxiliary supervisions are imposed.

$\diamond$ We demonstrate the value of face de-spoofing by its contribution to face anti-spoofing and the visualization of the spoof noise patterns.

\Section{Prior Work}
\figvspace
We review the most relevant prior works to ours from two perspectives: texture-based face anti-spoofing and de-X problems. 

%\Paragraph{Cue-based Methods:} some approaches attempt to leverage the spontaneous face motions. Eye-blinking is one cue proposed in [], to detect spoof attacks such as paper attack. In [], Kollreider et al. use lip motion to monitor the face liveness. Methods proposed in [] combine audio and visual cues to verify the face liveness.

\Paragraph{Texture-based Face Anti-spoofing} 
Texture analysis is widely adopted in face anti-spoofing as well as other computer vision tasks~\cite{krizhevsky2012imagenet,pose-invariant-face-alignment-via-cnn-based-dense-3d-model-fitting}, where defining an effective feature representation is the key endeavor. 
%Finding an effective feature representation for the spoof texture (i.e., spoof noise pattern) is a key solution to face anti-spoofing. 
Early works apply the hand-crafted feature descriptors, such as LBP~\cite{de2012lbp,de2013can,maatta2011face}, HoG~\cite{komulainen2013context,yang2013face}, SIFT~\cite{patel2016secure} and SURF~\cite{boulkenafet2017face}, to project the faces to a low-dimension embedding.
However, those hand-crafted features are not specifically designed to capture the subtle differences in the spoofing faces, and thus the embedding may not be discriminative. 
In addition, those features may not be robust to variations such as illumination, pose, and etc.
To overcome some of these difficulties, researchers tackle the problem in different domains, such as HSV and YCbCr color space~\cite{boulkenafet2015face,boulkenafet2016face}, temporal domain~\cite{siddiqui2016face,bao2009liveness,feng2016integration,xu2015learning} and Fourier spectrum~\cite{li2004live}. 

Heading into the deep learning era, researchers aim to build deep models for a higher accuracy. 
Most of the CNN works treat face anti-spoofing as a binary classification problem and apply the softmax loss function. 
Compared to hand-crafted features, such models~\cite{xu2015learning} achieve remarkable improvements in the intra-testing (i.e., train and test within the same dataset). However, during the cross-testing (i.e., train and test in different datasets), these CNN models exhibit a poor generalization ability due to the overfitting to training data.
Atoum et al.~\cite{atoum2017face} and Liu et al.~\cite{ learning-deep-models-for-face-anti-spoofing-binary-or-auxiliary-supervision} observe the overfitting issue of the softmax loss, and both propose novel auxiliary-driven loss functions instead of softmax to supervise the CNN. 
These works bring us the insight that we need to involve the domain knowledge to solve face anti-spoofing.

To the best of our knowledge, all the previous methods are discriminative models. 
There are only a few papers~\cite{patel2016secure,patel2016cross} trying to categorize the types and properties of the spoof noise pattern, such as color distortion and moir\'e pattern. %FIXME what are the difference to ours? Spell out clearly.
In this work, we analyze the properties of spoof noise and design a GAN-fashion generative model~\cite{disentangled-representation-learning-gan-for-pose-invariant-face-recognition} to estimate the spoof noise pattern and peel it off the spoof image.
We believe by decomposing the spoof image, CNN can analyze the spoof noise more directly and effectively, and gain more knowledge in tackling face anti-spoofing.

\Paragraph{De-X problems} 
De-X problems, such as de-noising, de-blurring, de-mosaicing, super-resolution and inpainting~\cite{buades2005non,lefkimmiatis2017non,taimemnet,tai2017image,zhang2017image,zhang2018densely,zhou2018deep,gharbi2016deep,criminisi2004region,bertalmio2003simultaneous,fsrnet-end-to-end-learning-face-super-resolution-with-facial-priors}, are classic low-level vision problems that remove the degradation effect or artifacts from the image.
General de-noising works assume additive Gaussian noise and researchers propose non-local filters~\cite{buades2005non} or CNNs~\cite{lefkimmiatis2017non,zhang2017image} to exploit the inherent similarity within the images.
For de-mosaicing and super-resolution, many models, such as ResNet in~\cite{taimemnet,tai2017image} and joint models in~\cite{zhou2018deep,gharbi2016deep,zhang2018densely}, are learnt from the given pairs of low-quality input and high-quality ground truth.
In image inpainting,  users mark the area to inpaint in a mask map and apply the filling based on the existing patch texture and the overall view structure in the unmasked region~\cite{liu2015comparison,criminisi2004region,bertalmio2003simultaneous}.

One advantage of existing de-X problems is that most of the image degradation can be easily synthesized. 
This brings two benefits: 1) it provides the model training with the input degraded samples and {\it golden} ground-truth original images for supervision. 2) it is easy to synthesize a large amount of data for training and evaluation. 
On the contrary, degradation due to spoofing is versatile, complex, and subtle. 
It consists of $2$-stage degradation: one from the spoof medium (e.g., paper and digital screen), and the other from the interaction of the spoof medium with the imaging environment. 
Each stage includes a large number of variations, such as medium type, illumination, non-rigid deformation and sensor types. 
Combination of these variations makes the overall degradation varies greatly. 
As a result, it is almost impossible to mimic realistic spoofing by synthesizing a degradation, which is a distinct challenge of face de-spoofing compared to the conventional de-X problems.

Without the ground truth of the degraded image, face de-spoofing becomes a very challenging problem. % and needs proper designs to tackle it.
In this work, we propose an encoder-decoder architecture with novel loss functions and supervisions to solve the de-spoofing problem.

\Section{Face De-spoofing}
\figvspace
In this section, we start with a case study of spoof noise pattern, which demonstrates a few important characteristics of the noise. 
This study motivates us to design the novel CNN architecture that will be presented in Sec.~\ref{sec_network_architecture}.

\SubSection{A Case Study of Spoof Noise Pattern}\label{case_study}

The core task of face de-spoofing is to estimate the spoofing-relevant noise pattern in the given face image.
Despite the strength of using a CNN model, we are still facing the challenge of learning {\it without} the ground truth of the noise pattern.
To address this challenge, we would like to first carry out a case study on the noise pattern with the objectives of answering the following questions: 1) is Eqn.~\ref{eqn_noisemodeling} a good modeling of the spoof noise? 2) what characteristics does the spoof noise hold?

\begin{figure*}[t!]
	\centering
	\small
	\includegraphics[width=\linewidth]{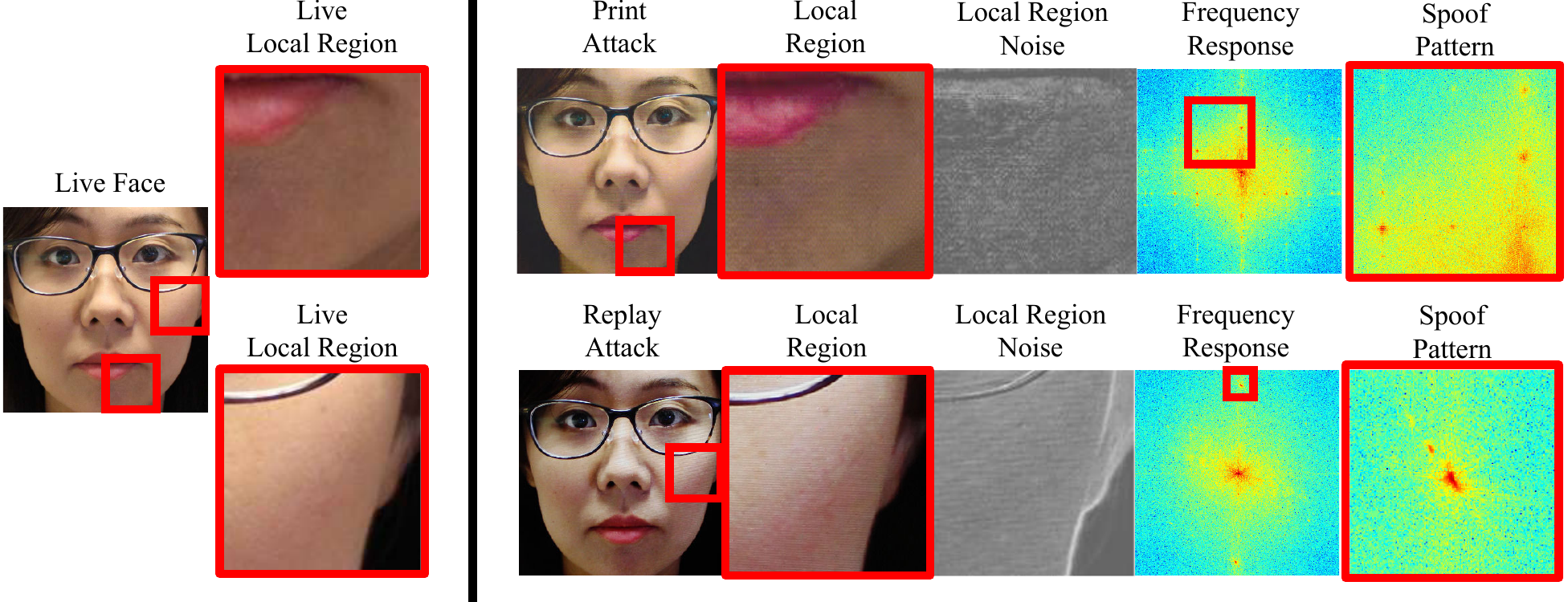} 
	\caption{\small The illustration of the spoof noise pattern. $\textbf{Left:}$ live face and its local regions. $\textbf{Right:}$ Two registered spoofing faces from print attack and replay attack. For each sample, we show the local region of the face, intensity difference to the live image, magnitude of $2$D FFT, and the local peaks in the frequency domain that indicates the spoof noise pattern. Best viewed electronically.} 
	\label{fig:noise}
	\figvspace
\end{figure*}

Let us denote a genuine face as $\mathbf{\hat{I}}$. 
By using printed paper or video replay on digital devices, the attacker can manufacture a spoof image  $\mathbf{I}$ from $\mathbf{\hat{I}}$. 
Considering no non-rigid deformation between $\mathbf{I}$ and $\mathbf{\hat{I}}$,
we summarize the degradation from  $\mathbf{\hat{I}}$ to $\mathbf{I}$ as the following steps:
\begin{enumerate}
 %\item Face $\mathbf{I_0(x,y)}$, being captured by camera, is discretized $\mathbf{x,y}$ coordinates and intensity. It creates pixel lattice and generate digitalized live image:
%\begin{equation}
%\label{eq:d1}
%\begin{split}
%\mathbf{I(x,y) = [I_0(x,y) + n_{lattice}(x,y)]}  \\
%\mathbf{n_{lattice}(x,y) = - I_0(x,y),  \text{when } x,y \notin \mathbb{Z}}
%\end{split}
%\end{equation}
\item \textbf{Color distortion:} Color distortion is due to a narrower color gamut of the spoof medium (e.g. LCD screen or Toner Cartridge). %, and camera (from the face recognition system) auto white balancing based on the environment lighting.  
It is a projection from the original color space to a tinier color subspace. 
%It would not only shift the global tones of $\mathbf{I}$, but also cause high-frequency information loss. 
This noise is dependent on the color intensity of the subject, and hence it may apply as a degradation matrix to the genuine face $\mathbf{I}$ during the degradation.

\item \textbf{Display artifacts:} 
Spoof mediums often use several nearby dots/sensors to approximate one pixel's color, and they may also display the face differently than the original size. 
Approximation and down-sampling procedure would cause a certain degree of high-frequency information loss, blurring, and pixel perturbation. 
%In replay, people use pixels per inch (PPI) to describe the quality; and in image printing, people use dots per inch (DPI) to describe the quality. 
This noise may also apply as a degradation matrix due to its subject dependence.

\item \textbf{Presenting artifacts:} When presenting the spoof medium to the camera, the medium interacts with the environment and brings several artifacts, including reflection and transparency of the surface. 
This noise may apply as an additive noise.

\item \textbf{Imaging artifacts:}  Imaging lattice patterns such as screen pixels on the camera's sensor array (e.g. CMOS and CCD) would cause interference of light. 
This effect leads to aliasing and creates moir\'e pattern, which appears in replay attack and some print attack with strong lattice artifacts. %commonly
This noise may apply as an additive noise.
\end{enumerate}

\begin{figure*}[t]
	\centering
	\small
	\includegraphics[width=0.8\linewidth]{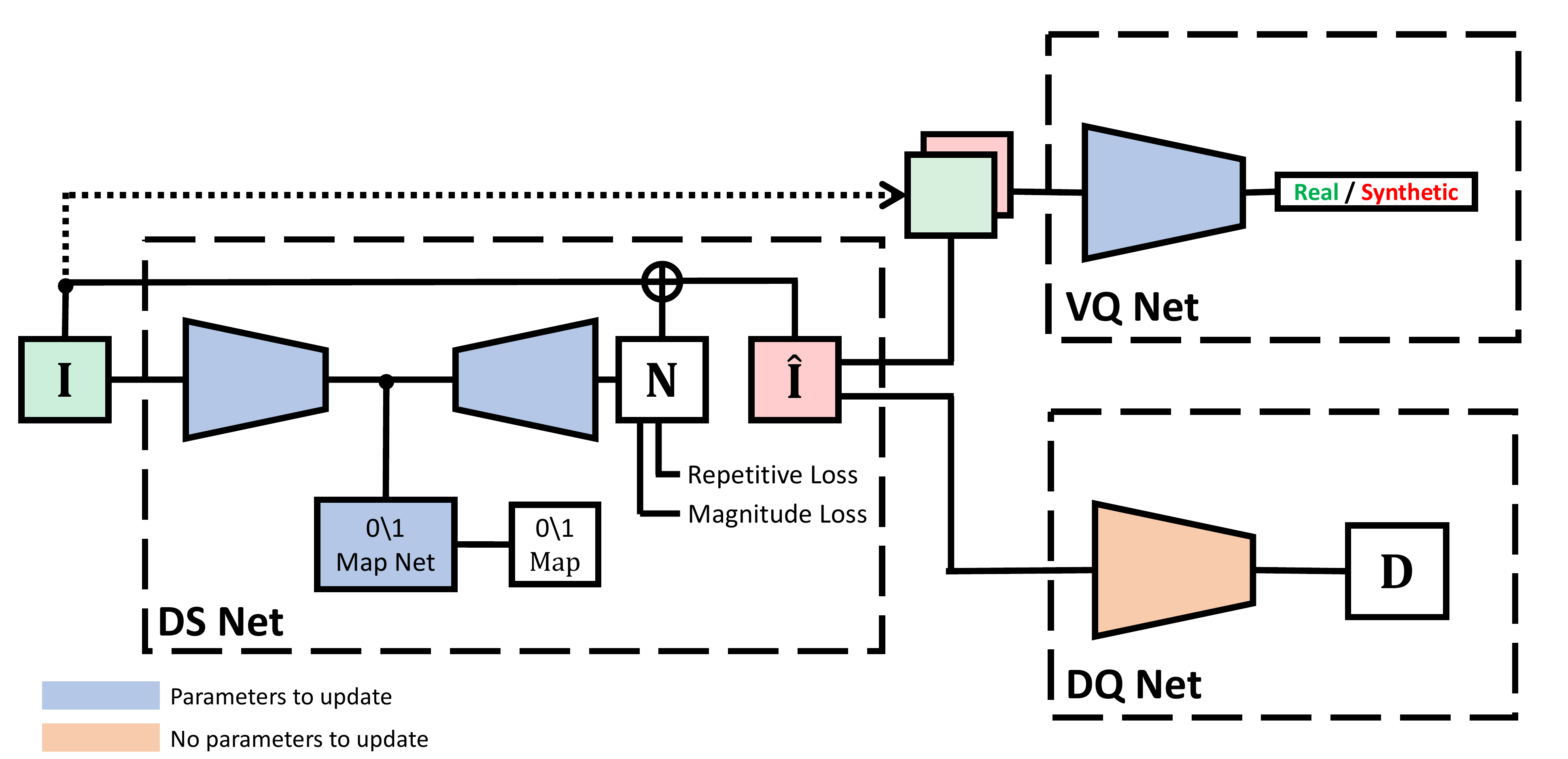} 
	\caption{\small The proposed network architecture.}
	\label{fig:architecture}
	\figvspace
\end{figure*}

These four steps show that the spoof image $\mathbf{I}$ can be generated via applying degradation matrices and additive noises to $\mathbf{\hat{I}}$, which is basically conveyed by Eqn.~\ref{eqn_noisemodeling}.
As expressed by Eqn.~\ref{eqn_noisemodeling}, the spoof image is the summation of the live image and image-dependent noise.
To further validate this model, we show an example in Fig.~\ref{fig:noise}. 
Given a high-quality live image, we carefully produce two spoof images via print and replay attack, with minimal non-rigid deformation.
After  each spoof image is registered with the live image, the live image becomes the {\it ground truth} live image if we would perform de-spoofing on the spoof image.
This allows us to compute the difference between the live and spoof images, which is the noise pattern $N(\mathbf{\hat{I}})$.
% The spoof images are registered with the live image so that the subtraction of live and spoof images can present the spoof noise. 
To analyze its frequency properties, we perform FFT on the spoof noise and show the $2$D shifted magnitude response.
 
In both spoof cases, we observe a high response in the low-frequency domain, which is related to color distortion and display artifacts.
In print attack, \textit{repetitive} noise in Step $3$ leads to a few ``peak"  responses in the high-frequency domain. % and globally existing noise pattern accumulate the same response in the frequency domain, making them the peaks.
Similarly, in the replay attack, visible moir\'e pattern reflects as several spurs in the low-frequency domain, and the lattice pattern that causes the moir\'e pattern is represented as peaks in the high-frequency domain.
Moreover, spoof patterns are uniformly distributed in the image domain  due to the uniform texture of the spoof mediums. And the high response of the repetitive pattern in the frequency domain exactly demonstrates that it appears widely in the image and thus can be viewed as ubiquitous.
%To note that, in practice, it's hard to register the spoof face with the live image due to unknown deformation of the face (e.g. warping or tilting the spoof medium). 

Under this ideal registration, the comparison between live and spoof images provides us a basic understanding of the spoof noise pattern. 
It is a type of texture that has the characteristics of $\textbf{repetitive}$ and $\textbf{ubiquitous}$. 
%Combining the above steps of degradation, the overall spoof degradation can be represented as Equation~\ref{eq:Model}. 
Based on this modeling and noise characteristics, we design a network to estimate the noise {\it without} the access to the precisely registered ground truth live image, as this case study has.
%Based on this knowledge, we propose our model architecture and several novel loss functions to specifically decompose the spoof noise without registering and pairing the live face and spoof face. 

%\SubSection{Network Overview}
%\label{sec_network_architecture}

\figvspace
\SubSection{De-Spoof Network}

\subsubsection{Network Overview:}\label{sec_network_architecture}

\begin{table}[t!]
\caption{\small The network structure of DS Net, DQ Net and VQ Net. Each convolutional layer is followed by an exponential linear unit (ELU) and batch normalization layer.  The input image size for DS Net is $256\times 256\times 6$. All the convolutional filters are $3\times 3$. 0\textbackslash 1 Map Net is the  bottom-left part, i.e., conv1-10, conv1-11, and conv1-12.} % FConv represents fractionally-strided convolution.}
\label{tab:network}
\figvspace 
\begin{center}
\small
\resizebox{0.99\linewidth}{!}{
\setlength{\tabcolsep}{3pt}
\begin{tabular}{ cccc||ccccc||ccccc||cccc }
%\toprule
\hline\hline%
\multicolumn{3}{c}{DS Net (Encoder Part)} 
& \hspace{1mm} & \hspace{1mm} 
& \multicolumn{3}{c}{DS Net (Decoder Part)} 
& \hspace{1mm} & \hspace{1mm} 
& \multicolumn{3}{c}{DQ Net}
& \hspace{1mm} & \hspace{1mm}  
& \multicolumn{3}{c}{VQ Net}\\
%\cmidrule(r){1-3}
%\cmidrule(r){5-7}

Layer & Chan./Stri. & Outp. Size & & &
Layer & Chan./Stri. & Outp. Size & & & 
Layer & Chan./Stri. & Outp. Size & & &
Layer & Chan./Stri. & Outp. Size
\\ \hline%\midrule

%input
\multicolumn{3}{c}{\textbf{Input}} &&& \multicolumn{3}{c}{\textbf{Input}} &&& \multicolumn{3}{c}{\textbf{Input}} &&& \multicolumn{3}{c}{\textbf{Input}}\\
\multicolumn{3}{c}{image} &&& \multicolumn{3}{c}{pool1-1+pool1-2+pool1-3} &&& \multicolumn{3}{c}{\{image,live\}} &&& \multicolumn{3}{c}{\{image,live\}}\\
\hline%\midrule

conv1-0 & 24/1 & 256 &&& resize & -/- & 256 &&& conv3-0 & 64/1 & 256 &&&  &  & \\

\hline
conv1-1 & 20/1 & 256 &&& conv2-1 & 28/1 & 256 &&& conv3-1 & 128/1 & 256 &&& conv4-1 & 24/2 & 256\\
conv1-2& 25/1 & 256 &&& conv2-2 & 24/1 & 256 &&& conv3-2 & 196/1 & 256 &&& conv4-2 & 20/2 & 256\\
conv1-3 & 20/1 & 256 &&&&&&&& conv3-3 & 128/1 & 256 &&&pool4-1  & -/2 & 128\\
pool1-1  & -/2 & 128 &&&&&&&& pool3-1  & -/2 & 128 &&&&& \\

\hline
conv1-4 & 20/1 & 128 &&& conv2-3 & 20/1 & 256 &&& conv3-4 & 128/1 & 128 &&& conv4-3 & 20/1 & 128\\
conv1-5 & 25/1 & 128 &&& conv2-4 & 20/1 & 256 &&& conv3-5 & 196/1 & 128 &&& conv4-4 & 16/1 & 128\\
conv1-6 & 20/1 & 128 &&&&&&&& conv3-6 & 128/1 & 128 &&&pool4-2  & -/2 & 64\\
pool1-2  & -/2 & 64 &&&&&&&& pool3-2  & -/2 & 64 &&&&& \\

\hline
conv1-7 & 20/1 & 64 &&& conv2-5 & 20/1 & 256 &&& conv3-7 & 128/1 & 64 &&& conv4-5 & 12/1 & 64\\
conv1-8 & 25/1 & 64 &&& conv2-6 & 16/1 & 256 &&& conv3-8 & 196/1 & 64 &&& conv4-6 & 6/1 & 64\\
conv1-9 & 20/1 & 64 &&&&&&&& conv3-9 & 128/1 & 64 &&& pool4-3  & -/2 & 32\\
pool1-3 & -/2 & 32 &&&&&&&& pool3-3  & -/2 & 32 &&&&& \\

\hline
\multicolumn{3}{c}{\textbf{short-cut connection}} &&& \multicolumn{3}{c}{} &&& \multicolumn{3}{c}{\textbf{short-cut connection}} &&& \multicolumn{3}{c}{\textbf{vectorize}}\\
\multicolumn{3}{c}{pool1-1+pool1-2+pool1-3} &&& \multicolumn{3}{c}{} &&& \multicolumn{3}{c}{pool3-1+pool3-2+pool3-3} &&& & & 1024\\

\hline
conv1-10 & 28/1 & 32 &&& conv2-7 & 16/1 & 256 &&& conv3-10 & 128/1 & 32 &&& fc4-1 & 1/1 & 100\\
conv1-11 & 16/1 & 32 &&& conv2-8 & 6/1 & 256 &&& conv3-11 & 64/1 & 32 &&& dropout & - & 0.2\%\\
conv1-12 & 1/1 & 32 &&& live & \multicolumn{2}{c}{(image  -  conv2-8)} &&& conv3-12 & 1/1 & 32 &&& fc4-2 & 1/1 & 2\\

\hline \hline  \\ %\midrule
\end{tabular}}
\end{center}\figvspace\vspace{-3mm}
\end{table}

Figure~\ref{fig:architecture} shows the overall network architecture of our proposed method. 
It consists of three parts: De-Spoof Net (DS Net), Discriminative Quality Net (DQ Net), and Visual Quality Net (VQ Net). 
DS Net is designed to estimate the spoof noise pattern $\mathbf{N}$ (i.e. the output of $N(\mathbf{\hat{I}})$) from the input image $\mathbf{I}$.  
The live face $\mathbf{\hat{I}}$ then can be reconstructed by subtracting the estimated noise $\mathbf{N}$ from the input image $\mathbf{I}$. 
%For the spoof images, following Eqn.~\ref{eqn_noisemodeling}, the subtraction of the noise from the original spoof image will lead to a reconstructed live image (i.e., the de-spoofed image).
%For the live images, the spoof noise pattern $\mathbf{N}$ should be zero.
This reconstructed image $\mathbf{\hat{I}}$ should be both visually appealing and indeed live, which will be safeguarded by the DQ Net and VQ Net respectively. 
All networks can be trained in an end-to-end fashion. 
The details of the network structure are shown in Tab.~\ref{tab:network}.

%DS Net is designed to analyze the input image and decompose (or peel off) the spoof noise from the input image. 
As the core part, DS Net is designed as an encoder-decoder structure with the input $\mathbf{I}  \in \mathbb{R}^{256 \times 256 \times 6}$.
Here the $6$ channels are RGB $+$ HSV color space, following the suggestion in~\cite{atoum2017face}.  %and aims at detecting and deviding the spoof noise pattern from an input 
In the encoder part, we first stack $10$ convolutional layers with $3$ pooling layers.
%to extract the feature representation for each local reception field. 
Inspired by the residual network~\cite{he2016deep}, we follow by a short-cut connection: concatenating the responses from $pool1$-$1$, $pool1$-$2$ with $pool1$-$3$, and then sending them to $conv1$-$10$.
This operation helps us to pass the feature responses from different scales to the later stages and ease the training procedure.
%in the encoder to ease the training procedure.
Going through $3$ more convolution layers, the responses $\mathbf{F}  \in \mathbb{R}^{32 \times 32 \times 32}$ from $conv1$-$12$
are the feature representation of the spoof noise patterns.
%is the response of the filters that are designed to estimate spoof noise patterns. 
The higher magnitudes the responses have, the more spoofing-perceptible the input is.   

Out from the encoder, the feature representation $\mathbf{F}$ is fed into the decoder to reconstruct the spoof noise pattern. 
$\mathbf{F}$ is directly resized to the input spatial size $256\times 256$. It introduces no extra grid artifacts, which exist in the alternative approach of using a deconvolutional layer. 
%As shown in Fig.~\ref{fig:noise}, the spoof noise pattern is shown to be repetitive and ubiquitous textures. 
%Hence the feature map $\mathbf{F}$ can be resized to match the input/output spatial size, which avoids grid artifacts existing in the alternative approach of using deconvolutional layer. 
Then, we pass the resized $\mathbf{F}$ to several convolutional layers to reconstruct the noise pattern $\mathbf{N}$. 
According to Eqn.~\ref{eqn_noisemodeling}, the reconstructed live image can be retrieved by:
%\begin{equation}\label{eq:Model}
$\mathbf{\hat{x}} = \mathbf{x} - N(\mathbf{\hat{x}}) = \mathbf{I} - \mathbf{N}$.
%\end{equation}

Each convolutional layer in the DS Net is equipped with exponential linear unit (ELU) and batch normalization layers. 
To supervise the training of DS Net, we design multiple loss functions: losses from DQ Net and VQ Net for the image quality, 0\textbackslash 1 map loss, and noise property losses. 
We introduce these loss functions in Sec.~\ref{sec_gan}-\ref{sec_loss}.

\SubSection{DQ Net and VQ Net}
\label{sec_gan}
While we do not have the ground truth to supervise the estimated spoof noise pattern, it is possible to supervise the reconstructed live image, which implicitly guides the noise estimation.
To estimate a good-quality spoof noise, the reconstructed live image should be quantitatively and visually recognized as live.
%To evaluate the quality of the estimated spoof noise pattern, 
For this purpose, we propose two networks in our architecture: Discriminative Quality Net (DQ Net) and Visual Quality Net (VQ Net). 
The VQ Net aims to guarantee the reconstructed live face is photorealistic. 
The DQ Net is proposed to guarantee the reconstructed face would indeed be considered as live, based on the judgment of a pre-trained face anti-spoofing network. The details of our proposed architecture are shown in Tab.~\ref{tab:network}. 

\figvspace
\subsubsection{Discriminative Quality Net:} 
We follow the state-of-the-art network architecture of face anti-spoofing~\cite{learning-deep-models-for-face-anti-spoofing-binary-or-auxiliary-supervision} to build our DQ Net. %, shown in Tab.~\ref{}.  
It is a fully convolutional network with three filter blocks and three additional convolutional layers. 
Each block consists of three convolutional layers and one pooling layer. 
The feature maps after each pooling layer are resized and stacked to feed into the following convolutional layers. 
Finally, DQ Net is supervised to estimate the pseudo-depth $\mathbf{D}$ of an input face, where $\mathbf{D}$ for the live face is the depth of the face shape and $\mathbf{D}$ for the spoof face is a zero map as a flat surface. We adopt the 3D face alignment algorithm in~\cite{liu2017dense} to estimate the face shape and render the depth via Z-Buffering.

Similar to the previous work~\cite{johnson2016perceptual}, DQ Net is pre-trained to obtain the semantic knowledge of live faces and spoofing faces. And during the training of DS Net, the parameters of DQ Net are fixed. 
Since the reconstructed images $\mathbf{\hat{I}}$ are live images, the corresponding pseudo-depth $\mathbf{D}$ should be the depth of the face shape.
%We consider the ground truth for all estimated live images, which are fed to DQ Net, as true depth maps. 
The backpropagation of the error from DQ Net guides the DS Net to estimate the spoof noise pattern which should be subtracted from the input image,

\begin{equation}\label{eq:DQNet}
 J_{DQ} = \left\|\text{CNN}_{DQ}(\mathbf{\hat{I}} ) - \textbf{D} \right\|_1,
\end{equation}
where $\text{CNN}_{DQ}$ is a fixed network and $\textbf{D}$ is the depth of the face shape. % for both live and spoof images.

%We construct the loss function from the last layer in order to obtain the largest reception field. % FIXME why do this?
%The error is back-propogated from DQ Net to De-spoof Net while keeping the parameters of DQ Net to be fixed.

\figvspace
\subsubsection{Visual Quality Net:} 
We deploy a GAN to verify the visual quality of the estimated live image $\mathbf{\hat{I}}$. 
Given both the real live image $\mathbf{I_{live}}$ and the synthesized live image $\mathbf{\hat{I}}$, VQ Net is trained to distinguish between $\mathbf{I_{live}}$  and $\mathbf{\hat{I}}$. Meanwhile, DS Net tries to reconstruct photorealistic live images where the VQ Net would classify them as non-synthetic (or real) images. The VQ Net consists of $6$ convolutional layers and a fully connected layer with a $2$D vector as the output, which represents the probability of the input image to be real or synthetic. In each iteration during the training, the VQ Net is evaluated with two batches, in the first one, the DS Net is fixed and we update the VQ Net,
\begin{equation}\label{eq:VQNet_train}
 J_{VQ_{train}} = -\mathbb{E}_{\mathbf{I} \in \mathcal{R}} \ \text{log}(\text{CNN}_{VQ}(\mathbf{I}))-\mathbb{E}_{\mathbf{I} \in \mathcal{S}} \ \text{log}(1-\text{CNN}_{VQ}(\text{CNN}_{DS}(\mathbf{I}))),
\end{equation}
where $\mathcal{R}$ and $\mathcal{S}$ are the sets of real and synthetic images respectively. In the second batch, the VQ Net is fixed and the DS Net is updated,
\begin{equation}\label{eq:VQNet_test}
J_{VQ_{test}} = -\mathbb{E}_{\mathbf{I} \in \mathcal{S}} \ \text{log}(\text{CNN}_{VQ}(\text{CNN}_{DS}(\mathbf{I}))).
\end{equation}

\figvspace

\SubSection{Loss functions}
\label{sec_loss}
%The main challenge for the face de-spoofing problem is that there is no ground truth for the spoof noise pattern. 
The main challenge for spoof modeling is the lack of the ground truth for the spoof noise pattern. 
%With VQ Net and  DQ Net alone, the training of DS Net may trap into some difficulties of describing the subtle changes.
%Meanwhile, we have some assumptions about the spoof noise, which have been validated from the case study in Sec.~\ref{case_study}. 
Since we have concluded some properties about the spoof noise in Sec.~\ref{case_study}, we can leverage them to design several novel loss functions to constrain the convergence space.
First, we introduce magnitude loss to enforce the spoof noise of the live image to be zero. 
Second, zero\textbackslash one map loss is used to demonstrate the ubiquitousness of the spoof noise. Third, we encourage the repetitiveness property of spoof noise via repetitive loss.
%which is utilized for representing the noise has the similar effect to the different part of the image. 
%These assumptions enable us to design novel loss functions to constraint the learning and facilitate the training. 
We describe three loss functions as the following:

\figvspace
\subsubsection{Magnitude Loss:} 
The spoof noise pattern for the live images is zero. 
The magnitude loss can be utilized to impose the constraint for the estimated noise. 
Given the estimated noise $\mathbf{N}$ and reconstructed live image $\mathbf{\hat{I}}=\mathbf{I}-\mathbf{N}$ of an original live image $\mathbf{I}$, we have,    
\begin{equation}\label{eq:MagnitudeLoss}
J_m =\left\|\mathbf{N}\right\|_1.
\end{equation}
%the first term also guides the network to generate live images without artifacts form the original spoof image.  

\Paragraph{Zero\textbackslash One Map Loss:}
%We should train the encoder of the DS Net to focus on the small changes of the spoof noise pattern in the images, so that, the decoder of the DS net be able to generate the noise for converting the input image to the reconstructed live one. 
To learn discriminative features in the encoder layers, we define a sub-task in the DS Net to estimate a zero-map for the live faces and an one-map for the spoof. 
Since this is a per {\it pixel} supervision, it is also a constraint of ubiquitousness on the noise.
Moreover, 0\textbackslash 1 map enables the receptive field of each pixel to cover a local area, which helps to learn generalizable features for this problem.
Formally, given the extracted features $\mathbf{F}$ from an input face image $\mathbf{I}$ in the encoder, we have,
 \begin{equation}\label{eq:ZeroOneLoss}
 J_z = \left\|\text{CNN}_{01map}(\mathbf{F}; \Theta ) - \textbf{M} \right\|_1,
\end{equation}
where $\mathbf{M} \in \mathbf{0}^{32\times 32}$ or $\mathbf{M} \in \mathbf{1}^{32\times 32}$ is the zero\textbackslash one map label. 

\figvspace\vspace{-2mm}
\subsubsection{Repetitive Loss:} 
Based on the previous discussion, we assume the spoof noise pattern to be repetitive, because it is generated from the repetitive spoof medium.
%The spoof medium affects different parts of the image similarly, and leads to high-frequency noise pattern. 
%We utilize this assumption in the process of estimating the noise, by encouraging the repetitiveness of the estimated spoof noise to be repetitive. 
%We encourage the repetitiveness of the estimated spoof noise in the process of estimating the spoof noise. 
To encourage the repetitiveness, we convert the estimated noise $\mathbf{N}$ to the Fourier domain and compute the maximum value in the high-frequency band. The existence of high peak is indicative of the repetitive pattern.
We would like to maximize this peak for spoof images, but minimize it for live images, as the following loss function:

\[ J_r =
\left \{
  \begin{tabular}{cc}
  $-\text{max}(H(\mathcal{F}(\mathbf{N}),k))$, & $\mathbf{I} \in Spoof$  \\
  $\left\|\text{max}(H(\mathcal{F}(\mathbf{N}),k))\right\|_1$,  & $\mathbf{I} \in Live$
  \end{tabular}
\right. 
,\]
where $\mathcal{F}$ is the Fourier transform operator, $H$ is an operator for masking the low-frequency domain of an image, i.e., setting a $k\times k$ region in the center of the shifted $2$D Fourier response to zero. 
%FIXME what are the parameters of H? what are your choice? how do they affect the performance?

Finally, the total loss function in our training is the weighted summation of the aforementioned loss functions and the supervisions for the image qualities,
\begin{equation}\label{eq:TotalLoss}  
J_T = J_z + \lambda_1 J_m + \lambda_2 J_r + \lambda_3 J_{DQ} + \lambda_4 J_{VQ_{test}} ,  
\end{equation}
where $\lambda_1, \lambda_2, \lambda_3, \lambda_4$ are the weights. 
During the training, we alternate between optimizing Eqn.~\ref{eq:TotalLoss} and Eqn.~\ref{eq:VQNet_train}.
%FIXME
% what is the training strategy? what optimizor do we use? etc. more details!!!
\Section{Experimental Results}

\SubSection{Experimental Setup}
\Paragraph{Databases} We evaluate our work on three face anti-spoofing databases, with print and replay attacks: Oulu-NPU~\cite{OULU_NPU_2017}, CASIA-MFSD~\cite{zhang2012face} and Replay-Attack~\cite{Chingovska_BIOSIG-2012}. Oulu-NPU~\cite{OULU_NPU_2017} is a high-resolution database, considering many real-world variations. 
Oulu-NPU also includes $4$ testing protocols:
Protocol $1$ evaluates on the illumination variation, Protocol $2$ examines the influence of different spoof medium, Protocol $3$ inspects the effect of different camera devices and Protocol $4$ contains all the challenges above, which is close to the scenario of cross testing.
CASIA-MFSD~\cite{zhang2012face} contains videos with resolution $640\times 480$ and $1280\times 720$. Replay-Attack~\cite{Chingovska_BIOSIG-2012} includes videos of $320\times 240$.
These two databases  are often used for cross testing~\cite{patel2016cross}.

\Paragraph{Parameter setting} We implement our method in Tensorflow~\cite{tensorflow2015-whitepaper}. Models are trained with the batch size of $6$ and the learning rate of $3\mathrm{e}{-5}$. We set the $k=64$ in the repetitive loss and set $\lambda_1$ to $\lambda_4$ in Eqn.~\ref{eq:TotalLoss} as $3, 0.005, 0.1$ and $0.016$, respectively. DQ Net is trained separately and remains fixed during the update of DS Net and VQ Net, but all sub-networks are trained with the same and respective data in each protocol.

\Paragraph{Evaluation metrics} To compare with previous methods, we use Attack Presentation Classification Error Rate (\textit{APCER})~\cite{acer1}, Bona Fide Presentation Classification Error Rate (\textit{BPCER})~\cite{acer1} and, $\textit{ACER}=(\textit{APCER}+\textit{BPCER})/2$~\cite{acer1}
for the intra testing on Oulu-NPU, and Half Total Error Rate (\textit{HTER})~\cite{bengio2004statistical}, half of the
summation of FAR and FRR, for the cross testing between CASIA-MFSD and Replay-Attack.

\SubSection{Ablation Study} 
Using Oulu-NPU Protocol $1$, we perform three studies on the effect of score fusing, the importance of each loss function, and the influence of image resolution and blurriness. %For experiments in the ablation study, we analyze on Oulu-NPU Protocol $1$.

\begin{table}[t!]
\small
	\centering
	\caption{The accuracy of different outputs of the proposed architecture and their fusions.}
	\resizebox{0.95\textwidth}{!} 
{
	\begin{tabular}{|c|c|c|c|c|c|c|c|}
		\hline
\multirow{2}{*}{Method} &\multirow{2}{*}{$0$\textbackslash$1$ map} & \multirow{2}{*}{Spoof noise} & \multirow{2}{*}{Depth map} &  \multicolumn{2}{c|}{Fusion (Spoof noise, Depth map)}&  \multicolumn{2}{c|}{Fusion of all three outputs} \\ \cline{5-8}  
&&&& Maximum & Average& Maximum & Average  \\\hline
APCER &$2.50$&$1.70$&$1.66$&$1.70$&$1.27$&$1.70$&$1.27$ \\\hline  
BPCER &$2.52$&$1.70$&$1.68$&$1.73$&$1.73$&$1.73$&$1.73$ \\\hline  
ACER &$2.51$&$1.70$&$1.67$&$1.72$&$1.50$&$1.72$&$1.50$\\\hline  
	\end{tabular}
	} 
\label{tab:fusion}
\figvspace
\end{table}

\Paragraph{Different fusion methods} In the proposed architecture, three outputs can be utilized for classification: the norms of either the 0\textbackslash 1 map, the spoof noise pattern or the depth map. 
Because of the discriminativeness enabled by our learning, we can simply use a rudimentary classifier like $L$-$1$ norm.
Note that a more advance classifier is applicable and would likely lead to higher performance.
Table~\ref{tab:fusion} shows the performance of each output and their fusion with maximum and average. 
It shows that the fusion of spoof noise and depth map achieves the best performance. However, adding the 0\textbackslash 1 map scores do not improve the accuracy since it contains the same information as the spoof noise. 
Hence, for the rest of experiments, we report performance from the average fusion of the spoof noise $\mathbf{N}$ and the depth map $\mathbf{\hat{D}}$, i.e., $score =(\left\|\mathbf{N}\right\|_1+\left\|\mathbf{\hat{D}}\right\|_1)/2$.

%\begin{equation}\label{eq:Fusion}
%score =(\left\|\mathbf{N}\right\|_1+\left\|\mathbf{\hat{D}}\right\|_1)/2,
%\end{equation}

\Paragraph{Advantage of each loss function} We have three main loss functions in our proposed architecture. 
To shows the effect of each loss function, we train a network with each loss excluded one by one. 
By disabling the magnitude loss, the $0$\textbackslash$1$ map loss and the repetitive loss, we obtain the ACERs $5.24$, $2.34$ and $1.50$, respectively. 
To further validate the repetitive loss, we perform an experiment on high-resolution images by changing the network input to the cheek region of the original $1080$P resolution.
The ACER of the network with the repetitive loss is $2.92$ but
the network without cannot converge. 
%These performances show the importance of the $0$\textbackslash$1$ map and the magnitude losses for guiding the network.

\begin{table}[t!]
%\figvspace
\small
	\centering
	\caption{\small ACER of the proposed method with different image resolutions and blurriness. To create blurry images, we apply Gaussian filters with different kernel sizes to the input images.}
\begin{tabular}{ccc}
	\scalebox{0.8}{
	\begin{tabular}{|c|c|c|c|}
		\hline
		\backslashbox{Metric}{Resolution}& $256\times 256$ & $128\times 128$ & $64\times 64$\\ \hline		
		APCER & $1.27$ & $2.27$ & $5.24$ \\ \hline
		BPCER & $1.73$ & $3.36$ & $5.30$ \\ \hline
		ACER & $1.50$ & $3.07$ & $5.27$ \\ \hline
	\end{tabular}
	}
	
	& \qquad &	
	 
	\scalebox{0.8}{
	\begin{tabular}{|c|c|c|c|c|c|}
		\hline
		\backslashbox{Metric}{Blurriness}& $1\times 1$ & $3\times 3$ & $5\times 5$& $7\times 7$& $9\times 9$\\ \hline		
		APCER & $1.27$ & $2.29$ & $3.12$ & $3.95$& $4.79$\\ \hline
		BPCER & $1.73$ & $2.50$ & $3.33$ & $4.16$& $5.00$\\ \hline
		ACER & $1.50$ & $2.39$ & $3.22$ & $4.06$& $4.89$\\ \hline
	\end{tabular}
	} 
\end{tabular}
\label{tab:resolution}
\figvspace
\end{table}

\Paragraph{Resolution and blurriness} As shown in the ablation study of repetitive loss, the image quality is critical for achieving a high accuracy.  The spoof noise pattern may not be detected in the low-resolution or motion-blurred images. The testing results on different image resolutions and blurriness are shown in Tab.~\ref{tab:resolution}.  These results validate that the spoof noise pattern is less discriminative for the lower-resolution or blurry images, as the high-frequency part of the input images contains most of the spoof noise pattern.

%The $ACER$ of input images with resolution of $256\times 256$, $128\times 128$ and $64\times 64$ are $1.5$, $3.0$ and $5.2$ respectively.

%\textvspace
\SubSection{Experimental Comparison}
To show the performance of our proposed method, we present our accuracy in the intra testing of Oulu-NPU and the cross testing on CASIA and Replay-Attack. 
\textvspace
\begin{table}[t!]
\small
	\centering
	\caption{The intra testing results on $4$ protocols of Oulu-NPU.}
	\resizebox{0.7\textwidth}{!} 
{
	\begin{tabular}{|c|c|c|c|c|}
		\hline
Protocol &Method & APCER (\%) & BPCER (\%) & ACER (\%)\\ \hline  \hline
    & CPqD\cite{boulkenafet17ijcb} & $2.9$ & $10.8$ & $6.9$\\ \cline{2-5}
$1$ & GRADIANT\cite{boulkenafet17ijcb}& $1.3$ & $12.5$ & $6.9$\\ \cline{2-5}
    & Auxiliary\cite{learning-deep-models-for-face-anti-spoofing-binary-or-auxiliary-supervision}& $1.6$ & $\textbf{1.6}$  & $1.6$\\ \cline{2-5}
    & Ours & $\textbf{1.2}$ & $1.7$& $\textbf{1.5}$\\ \hline \hline
    & MixedFASNet\cite{boulkenafet17ijcb} & $9.7$ & $2.5$ & $6.1$\\ \cline{2-5}
$2$ & Ours & $4.2$ & $4.4$ & $4.3$\\ \cline{2-5}
    & Auxiliary\cite{learning-deep-models-for-face-anti-spoofing-binary-or-auxiliary-supervision} & $2.7$ & $2.7$ & $2.7$\\ \cline{2-5}
    & GRADIANT & $\textbf{3.1}$ & $\textbf{1.9}$ & $\textbf{2.5}$\\ \hline \hline
    &MixedFASNet & $5.3\pm6.7$ & $7.8\pm5.5$ & $6.5\pm4.6$\\  \cline{2-5}
$3$ &GRADIANT & $\textbf{2.6}\pm\textbf{3.9}$ & $5.0\pm5.3$ & $3.8\pm2.4$\\  \cline{2-5}
    & Ours  & $4.0\pm1.8$ & $3.8\pm1.2$ & $3.6\pm1.6$\\ \cline{2-5}
    & Auxiliary\cite{learning-deep-models-for-face-anti-spoofing-binary-or-auxiliary-supervision} & $2.7\pm1.3$ & $\textbf{3.1}\pm\textbf{1.7}$ & $\textbf{2.9}\pm\textbf{1.5}$\\ \hline \hline
    & Massy\_HNU~\cite{boulkenafet17ijcb} & $35.8\pm35.3$ & $8.3\pm4.1$ & $22.1\pm17.6$\\ \cline{2-5}
$4$ & GRADIANT & $\textbf{5.0}\pm\textbf{4.5}$ & $15.0\pm7.1$ & $10.0\pm5.0$\\ \cline{2-5}
    & Auxiliary\cite{learning-deep-models-for-face-anti-spoofing-binary-or-auxiliary-supervision} & $9.3\pm5.6$ & $10.4\pm6.0$ & $9.5\pm6.0$\\ \cline{2-5}
    & Ours  & $5.1\pm6.3$ & $\textbf{6.1}\pm\textbf{5.1}$ & $\textbf{5.6}\pm\textbf{5.7} $\\ \hline

	\end{tabular}
	} 
\label{tab:intraOulu}
\figvspace
\end{table}

\subsubsection{Intra Testing} We compare our intra testing performance on all $4$ protocols of Oulu-NPU. Table~\ref{tab:intraOulu} shows the comparison of our method and the best $3$ out of $18$ previous methods~\cite{learning-deep-models-for-face-anti-spoofing-binary-or-auxiliary-supervision,boulkenafet17ijcb}. 
Our proposed method achieves promising results on all protocols. 
Specifically, we outperform the previous state of the art by a large margin in Protocol $4$, which is the most challenging protocol, and similar to cross testing.

\textvspace
\subsubsection{Cross Testing} We perform cross testing between CASIA-MFSD~\cite{zhang2012face} and Replay-Attack~\cite{Chingovska_BIOSIG-2012}.
%to demonstrate the generalizability of the proposed method. 
As shown in Tab.~\ref{tab:cross}, our method achieves the competitive performance on the cross testing from CASIA-MFSD to Replay-Attack.
% by $5.94\%$. 
However, we achieve a worse HTER compared to the best performing methods from Replay Attack to CASIA-MFSD.  
We hypothesize the reason is that images of CASIA-MFSD are of much higher resolution than those of Replay Attack.
This shows that the model trained with higher-resolution data can generalize well on lower-resolution testing data, but not the other way around. 
This is one limitation of the proposed method, and worthy further research.

\begin{table}[t!]
\small
	\centering
	\caption{The HTER of different methods for the cross testing between the CASIA-MFSD and the Replay-Attack databases. We mark the top-$2$ performances in bold. }

	\begin{tabular}{|c|c|c|c|c|}
	
		\hline
\multirow{3}{*}{Method} & Train  & Test& Train & Test \\  \cline{2-5}
& CASIA & Replay& Replay & CASIA \\
& MFSD & Attack& Attack & MFSD \\ \hline
Motion~\cite{de2013can} & \multicolumn{2}{c}{$50.2$\%} & \multicolumn{2}{|c|}{$47.9$\%} \\  \hline
LBP-TOP~\cite{de2013can} & \multicolumn{2}{c}{$49.7$\%} & \multicolumn{2}{|c|}{$60.6$\%} \\  \hline
Motion-Mag~\cite{bharadwaj2013computationally} & \multicolumn{2}{c}{$50.1$\%} & \multicolumn{2}{|c|}{$47.0$\%} \\  \hline
Spectral cubes~\cite{pinto2015face} & \multicolumn{2}{c}{$34.4$\%} & \multicolumn{2}{|c|}{$50.0$\%} \\  \hline
CNN~\cite{yang2014learn} & \multicolumn{2}{c}{$48.5$\%} & \multicolumn{2}{|c|}{$45.5$\%} \\  \hline
LBP~\cite{boulkenafet2015face} & \multicolumn{2}{c}{$47.0$\%} & \multicolumn{2}{|c|}{$39.6$\%} \\  \hline
Colour Texture~\cite{boulkenafet2016face} & \multicolumn{2}{c}{$30.3$\%} & \multicolumn{2}{|c|}{$\textbf{37.7}$\%} \\  \hline
Auxiliary\cite{learning-deep-models-for-face-anti-spoofing-binary-or-auxiliary-supervision} & \multicolumn{2}{c}{$\textbf{27.6}$\%} & \multicolumn{2}{|c|}{$\textbf{28.4}$\%} \\  \hline
Ours & \multicolumn{2}{c}{$\textbf{28.5}$\%} & \multicolumn{2}{|c|}{$41.1$\%} \\  \hline
	\end{tabular}
\label{tab:cross}

\figvspace
\end{table}

\SubSection{Qualitative Experiments}

\subsubsection{Spoof medium classification} 
\begin{figure}[t!]
\small
	\centering
%	\resizebox{0.7\textwidth}{!} 
%{
	\begin{tabular}{ccc}
	\includegraphics[width=0.4\linewidth]{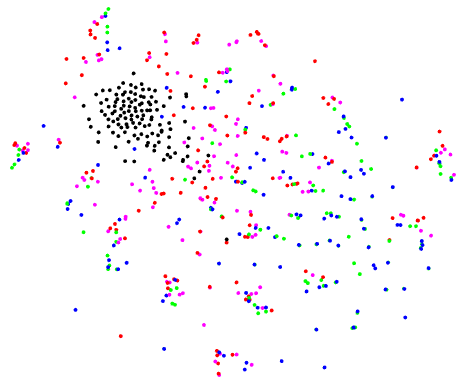} & \qquad \qquad & 
	\includegraphics[width=0.4\linewidth]{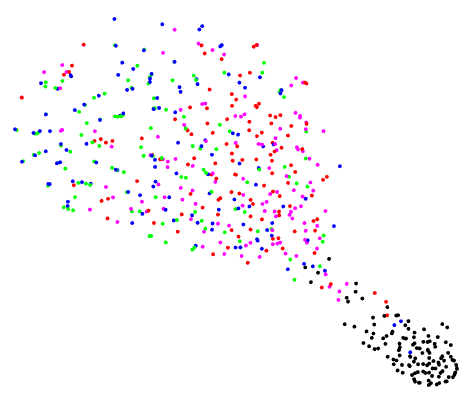} 
	\end{tabular}
%	} 
	\caption{The $2$D visualization of the estimated spoof noise for test videos on Oulu-NPU Protocol $1$. Left: the estimated noise, Right: the high-frequency band of the estimated noise, \textit{Color code} used: \textit{black}=live, \textit{green}=printer$1$, \textit{blue}=printer$2$, \textit{magenta}=display$1$, \textit{red}=display$2$.}
\label{fig:ClusterOulu}
\end{figure}

The estimated spoof noise pattern of the test images can be used for clustering them into different groups and each group represents
one spoof medium. 
To visualize the results, we use t-SNE~\cite{maaten2008visualizing} for dimension reduction. The t-SNE projects the noise $\mathbf{N}  \in \mathbb{R}^{256 \times 256 \times 6}$ to $2$ dimensions by best preserving the KL divergence distance. 
Fig.~\ref{fig:ClusterOulu} shows the distributions of the testing videos on Oulu-NPU Protocol $1$. 
The left image shows that the noise of live is well-clustered, and the noise of spoof is subject dependent, which is consistent with our noise assumption.
%even though the noise of the live videos are well-clustered, but . 
%As we believe the high frequency part is more subject independent, 
To obtain a better visualization, we utilize the high pass filter to extract the high-frequency information of noise pattern for dimension reduction. 
The right image shows that the high frequency part has more subject independent information about the spoof type and can be utilized for classification of the spoof medium.
%The right image shows a more separate distribution, due to the greater subject independence and spoof medium dependence of the high-frequency part.  

To further show the discriminative power of the estimated spoof noise, we divide the testing set of Protocol $1$ to training and testing parts and train an SVM classifier for spoof medium classification. 
We train two models, a three-class classifier (live, print and display) and a five-class classifier (live, print$1$, print$2$, display$1$ and display$2$), and they achieve the classification accuracy of $82.0\%$ and $54.3\%$ respectively, shown in Tab.~\ref{tab:confusionMatrix}. 
Most classification errors of the five-class model are within the same spoof medium. 
This result is noteworthy given that no label of spoof medium type is provided during the learning of the spoof noise model. 
Yet the estimated noise actually carries appreciable information regarding the medium type; hence we can observe reasonable results of spoof medium classification. 
This demonstrates that the estimated noise contains spoof medium information and indeed we are moving toward estimating the faithful spoof noise residing in each spoof image. 
In the future, if the performance of spoof medium classification improves, this could bring new impact to applications such as forensic.

\begin{table}[t!]
\small
	\centering
	\caption{\small The confusion matrices of spoof mediums classification based on spoof noise pattern.}
	\begin{tabular}{ccc}
	\scalebox{0.95}{
	\begin{tabular}{|c|c|c|c|}
		\hline
		\backslashbox{Actual}{Predicted}& live & print & display\\ \hline		
		live    & $59$ & $1$  & $0$ \\ \hline
		print   & $0$  & $88$ & $32$ \\ \hline
		display & $13$ & $8$  & $99$ \\ \hline
	\end{tabular}
	}
& \qquad & 
	\scalebox{0.95}{
	\begin{tabular}{|c|c|c|c|c|c|}
		\hline
		\backslashbox{Actual}{Predicted}& live & print$1$ & print$2$ & display$1$ & display$2$\\ \hline		
		live       & $59$ & $0$  & $1$  & $0$  & $0$ \\ \hline
		print$1$   & $0$  & $41$ & $2$  & $11$ & $6$ \\ \hline
		print$2$   & $0$  & $34$ & $11$ & $9$  & $6$ \\ \hline
		display$1$ & $10$ & $6$  & $0$  & $13$ & $31$ \\ \hline
		display$2$ & $8$  & $7$  & $0$  & $6$  & $39$ \\ \hline
	\end{tabular}
	}
	\end{tabular}
\label{tab:confusionMatrix}
\figvspace
\figvspace
\end{table}

\figvspace
\figvspace
\subsubsection{Successful and failure cases}
\begin{figure}[t!]
\small
	\centering
 
\begin{tabular}{c}
	\includegraphics[width=\linewidth]{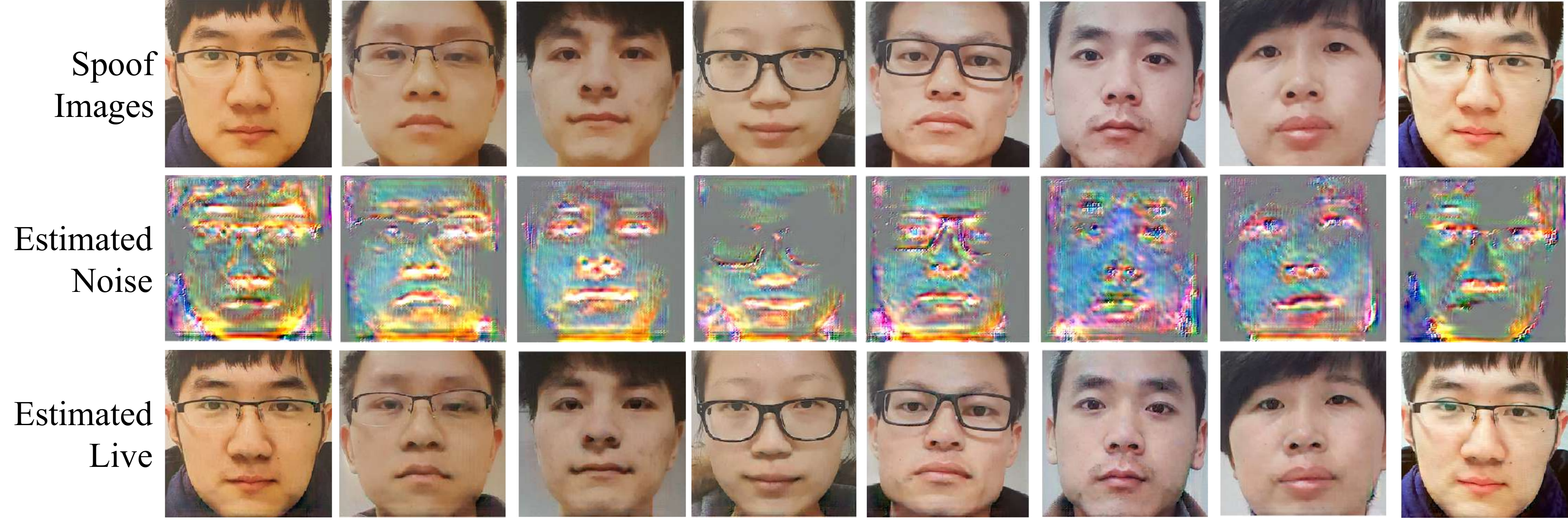} \\
   \includegraphics[width=\linewidth]{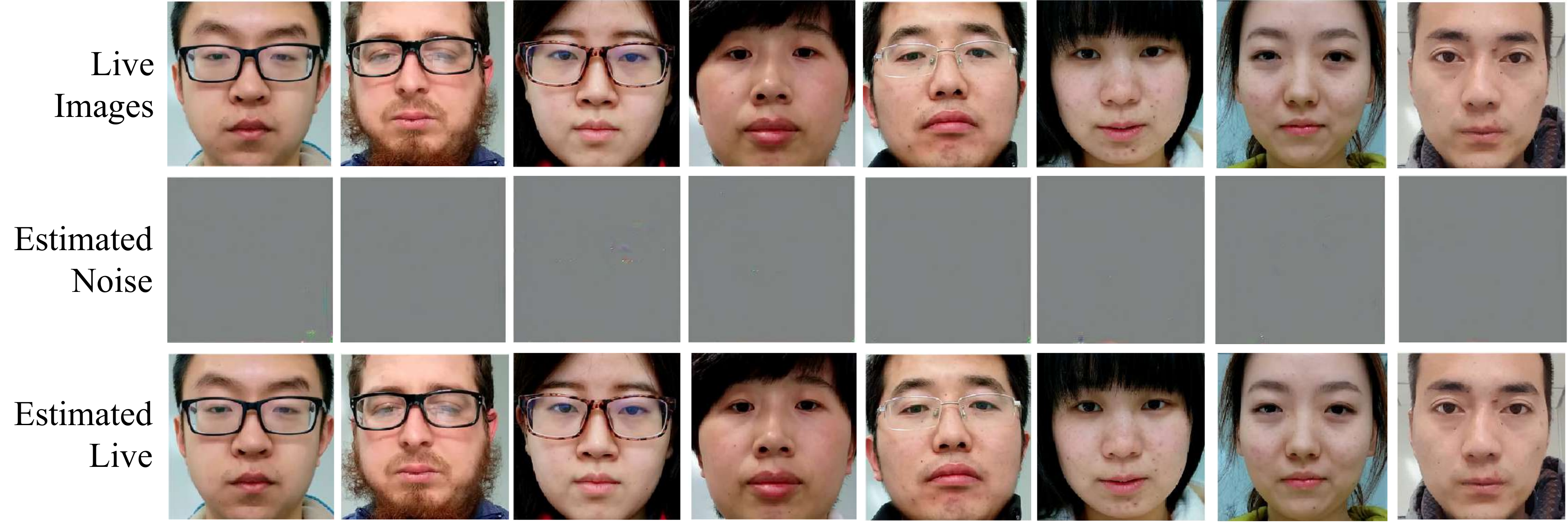} 
\end{tabular}
	\caption{The visualization of input images, estimated spoof noises and estimated live images for test videos of Protocol $1$ of Oulu-NPU database. The first four columns in the first row are paper attacks and the second four are the replay attacks. For a better visualization, we magnify the noise by $5$ times and add the value with $128$, to show both positive and negative noise.}
\label{fig:SuccessfulCases}
\end{figure}

\begin{figure}[t!]
\small
	\centering
 
	\includegraphics[width=\linewidth]{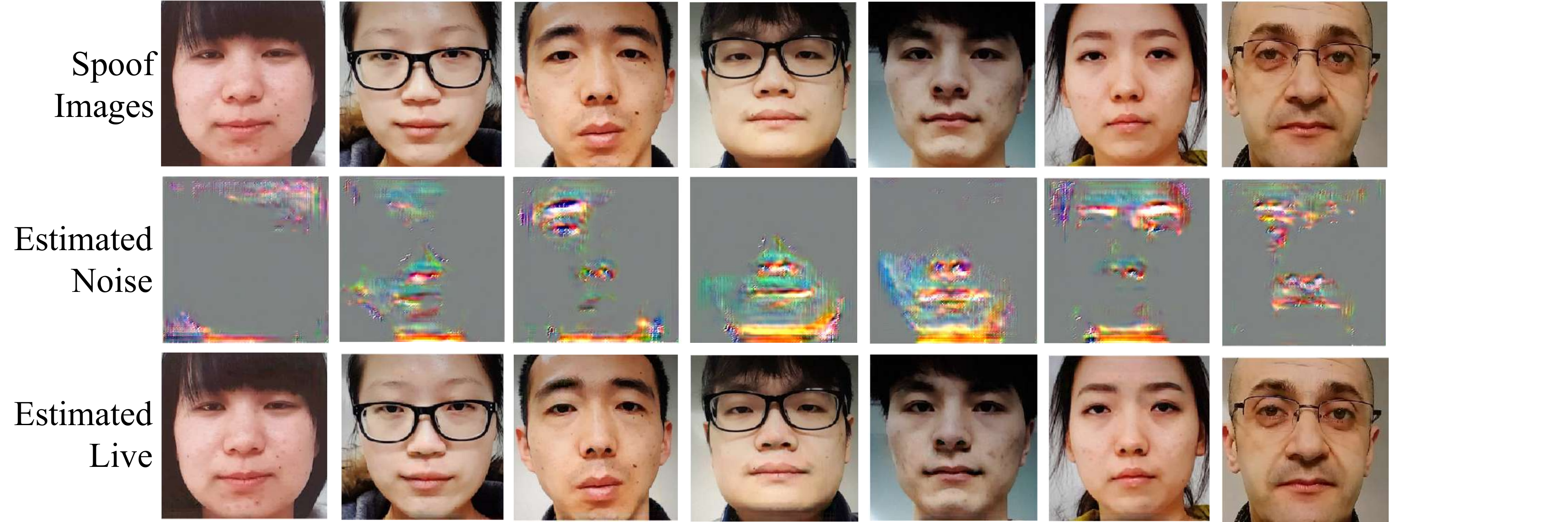} 

	\caption{The failure cases for converting the spoof images to the live ones.}
\label{fig:FailureCases}
\figvspace
\end{figure}

We show several success and failure cases in Fig.~\ref{fig:SuccessfulCases}-\ref{fig:FailureCases}. Fig.~\ref{fig:SuccessfulCases} shows that the estimated spoof noises are similar within each medium but different from the other mediums. We suspect that the yellowish color in the first four columns is due to the stronger color distortion in the paper attack. The fifth row shows that the estimated noise for the live images is nearly zero. For the failure cases, we only have a few false positive cases. The failures are due to undesired noise estimation which will motivate us for further research.

\Section{Conclusions}
This paper introduces a new perspective for solving the face anti-spoofing by inversely decomposing a spoof face into the live face and the spoof noise pattern. A novel CNN architecture with multiple appropriate supervisions is proposed. We design loss functions to encourage the pattern of the spoof images to be ubiquitous and repetitive, while the noise of the live images should be zero. We visualize the spoof noise pattern which can help to have a deeper understanding of the added noise by each spoof medium. We evaluate the proposed method on multiple widely-used face anti-spoofing databases.

\subsubsection{Acknowledgment}
This research is based upon work supported by the Office of the Director of National Intelligence (ODNI), Intelligence Advanced Research Projects Activity (IARPA), via IARPA R\&D Contract No.~$2017$-$17020200004$. The views and conclusions contained herein are those of the authors and should not be interpreted as necessarily representing the official policies or endorsements, either expressed or implied, of the ODNI, IARPA, or the U.S. Government. The U.S. Government is authorized to reproduce and distribute reprints for Governmental purposes notwithstanding any copyright annotation thereon.

\clearpage

\bibliographystyle{splncs}
\bibliography{abbrev_brief,egbib}
\end{document}